# A New Clustering neural network for Chinese word segmentation

Yuze Zhao

m.edward.zhao@gmail.com

## Abstract

In this article I proposed a new model to achieve Chinese word segmentation(CWS), which may have the potentiality to apply in other domains in the future. It is a new thinking in CWS compared to previous works: to consider it as a clustering problem instead of a labeling problem. In this model, LSTM and self-attention structures are used to collect context also sentence level features in every layer, and after several layers, a clustering model is applied to split characters into groups, which are the final segmentation results. I call this model CLNN. This algorithm can reach 98% of F-score (without OOV words) and 85%~95% F-score (with OOV words) in training data sets. Error analyses shows that OOV words will greatly reduce performances, which needs a deeper research in the future.

## 1   Introduction

Chinese word segmentation work has some difficult issues to resolve, mainly reflect in two aspects: Firstly, Characters may combine differently on various conditions, like"南京市长江大桥"may means a bridge in Nanjing, or the mayor of Nanjing. A huge number of algorithms aiming on it already have been developed, earlier solutions like Perceptron(Rosenblatt, 1958), CRF(Lafferty et al., 2001), or models based on N-gram(Brown et al., 1992). Lately with deep neural network ubiquitously applied in many scenarios, a growing number of structures have been raised in NLP and CWS domain. Li, Sun proposed a model based on punctuation marks (Li and Sun, 2009) and later they published THULAC tool. After LSTM been proposed in 1997 by Hochreater and Schmidhuber(Hochreiter and Schmidhuber, 1997), then advanced by Felix Gers(Gers, 2001) and Alex Graves(Graves et al., 2006) in 2001 and 2006, many researchers used this structure to solve context related issues. Huang, Xu and Yu first proposed BI-LSTM combined with CRF to do NLP benchmark sequence tagging(Huang et al., 2015). Luo, Xu et al. raised a custom neural network(Xu and Sun, 2016, Xu and Sun, 2017) combined with ADF algorithm(Sun et al., 2012), and got a high F-score of 90%-95% averagely, then they open-sourced it as PKUSEG(Luo et al., 2019). In 2017, Ashish Vaswani, Noam Shazeer et al. proposed an attention based encoder-decoder model Transformer(Vaswani et al., 2017) and then, Jacob Devlin, Ming-Wei Chang et al. raised Pre-training of Deep Bidirectional Transformers for Language Understanding, which is widely known as BERT(Devlin et al., 2018). This model is able to commendably resolve translate and labeling problems. In 2019, Huang, Cheng et al. advanced a model based on pre-trained BERT and CRF, with a new added projection layer and fine-tuning to adapt multiple domains(Huang et al., 2019).

Secondly, Word segmentation work may have "Jump Connection". Sentence "她一边吃饭一边看报" is segmented as "她/一边/吃饭/一边/看报" currently, but in fact, "一边……一边……"is a fixed phrase which distributed on discrete positions and could be grouped. Also, this phrase is commonly seen in English or other languages, like "so…that…" or "more…more…". Combine these words will help to understand the sentence better. From a higher perspective, we can conclude everything as vertexes in graph neural networks(Zhou et al., 2018), no pre-defined concepts like "verb" or "sentence". Some signals input into neural networks become vertexes to activate other vertexes which soon combine with other outer signals and then activate others again, this procedure is described in Figure 1. Models consisted of CRF only support segmentation in linear structure, but helpless in this scenario.

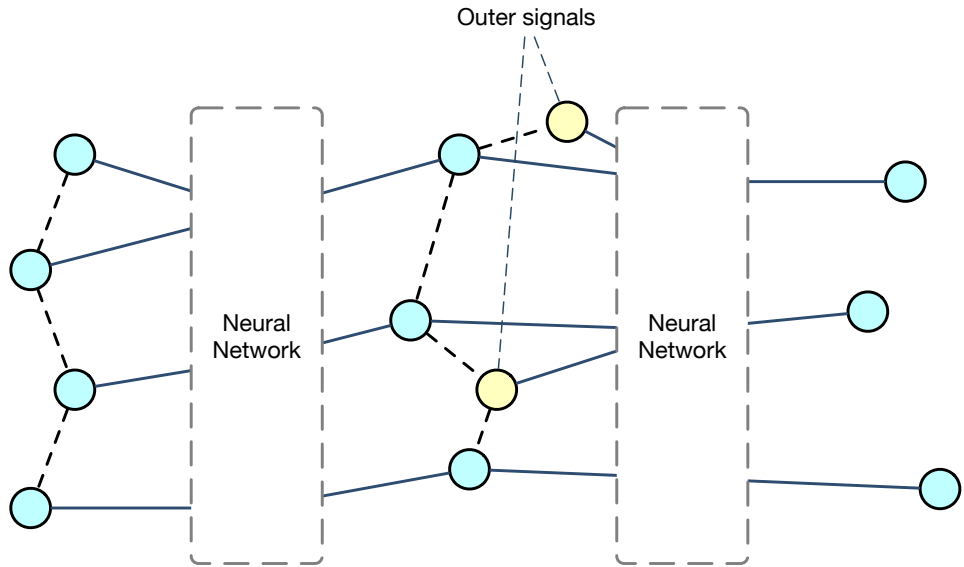

Figure 1: A simple show of vertexes activation.

Hence I propose a new model based on BI-LSTM and Self-Attention mechanism, and followed by a clustering model, which is able to split embedding into clusters to segment words beyond linear structure. The thinking behind this model is different from the models now exist: Treat CWS as a clustering problem, not a pure labeling problem.

## 2 Related Work

**BI-LSTM-CRF:** BI-LSTM-CRF(Huang et al., 2015) is the first model which combine LSTM with CRF and apply to NLP benchmark sequence labeling. Benefiting from LSTM, a wide range of context can be collected, mixed together and then abstracted to high-level features.
Finally, a CRF layer constraints labels with probabilities. Nevertheless, as described before, CRF model cannot recognize fixed phrase distributed on discrete positions.

**Transformer:** Attention Mechanism has a long story before transformer(Vaswani et al., 2017) proposed. Until 2017, Ashish Vaswani, Noam Shazeer et al used the Cartesian product of all embedding in a sentence and followed by a probability normalization to decide relations between words in encoder. Similarly, in the decoder they used the Cartesian product of intermediate hidden embedding with previous layer outputs to calculate new features of current layer, this mechanism constructs the skeleton of Transformer. The model uses multi-heads to get different relations between words and combine them together afterwards. Transformer resolves the long-dependency problem which RNN and LSTM cannot totally resolve. In this article, I also use self-attention mechanism to collect features and relations between characters, but simplify it to a single head.

## 3 The proposed clustering neural network

The structure of CLNN can be split into two parts: LSTM-Attention layers with Loss (supervised part), then a subsequent rectification and clustering model (unsupervised part). The overview can be viewed in Figure 2.

### 3.1 embedding

The first step is to build a vocabulary. All sentences in a data set will be parsed, all characters and words will be put into a vocabulary. This is explained in section 3.4. Transformer model suggests adding a positional embedding before feeding it into the attention-based models(Vaswani et al., 2017). Here I use the same algorithm as transformer did. Assuming embedding dimension is $d$ and sequence length is $n$:

$$X^d = \{x_1^d, x_2^d, x_3^d ... x_n^d\}$$
$$PE_{pos,2i} = \sin(pos/10000^{2i/d})$$
$$PE_{pos,2i+1} = \cos(pos/10000^{2i/d})$$
$$X_{pe} = X^d + PE$$

In which $X_{pe}$ will be the input of next model.

### 3.2 BI-LSTM sub layer

Given $X_{pe}$, it will be fed into a BI-LSTM sub layer which is able to collect context level features. This will help the model to understand when to split in different scenarios. Output dimension of BI-LSTM will be $2*d$, I use a linear model to map it to $d$:

$$h_{2d} = LSTM_{bi}(X_{pe})$$
$$Out_d = \omega \cdot h_{2d} + b$$

In which $LSTM_{bi}$ represents the non-linear procedure of BI-LSTM. This with the self-attention structure can be viewed in Figure 3.

## 3.3 Self-Attention sub layer

Embedding then enter a self-attention sub layer to combine related words after Res-Net(He et al., 2016) and Layer normalization(Ba et al., 2016). We assume the input as $In_d$:

$$q = W_q * In_d + b_q$$
$$k = W_k * In_d + b_k$$
$$v = W_v * In_d + b_v$$
$$Out_{atten} = \sum_{j=0}^{n} \alpha_{ij} * v$$
$$\alpha_i = \exp(q_i * k_j) \Big/ \sum_{j=0}^{n} \exp(q_i * k_j)$$

After self-attention sub layer, I add a $d$ to $N*d$ to $d$ feed-forward layer to extract features, with $N$ an expand ratio:

$$Out_{ff} = W_{ff}^d \big( ReLU(W_{ff}^{nd} * Out_{atten} + b_{ff}^{nd}) \big) + b_{ff}^d$$

Regularly, multiple layers can help to learn features better. I use six layers totally.

In this model, each sub layer has a Res-net(He et al., 2016) connection and Layer normalization(Ba et al., 2016) to prevent gradient disappearing and over-fitting. Dropout(Srivastava et al., 2014) rate can be set to 0.4 - 0.5, it depends on the data sets.

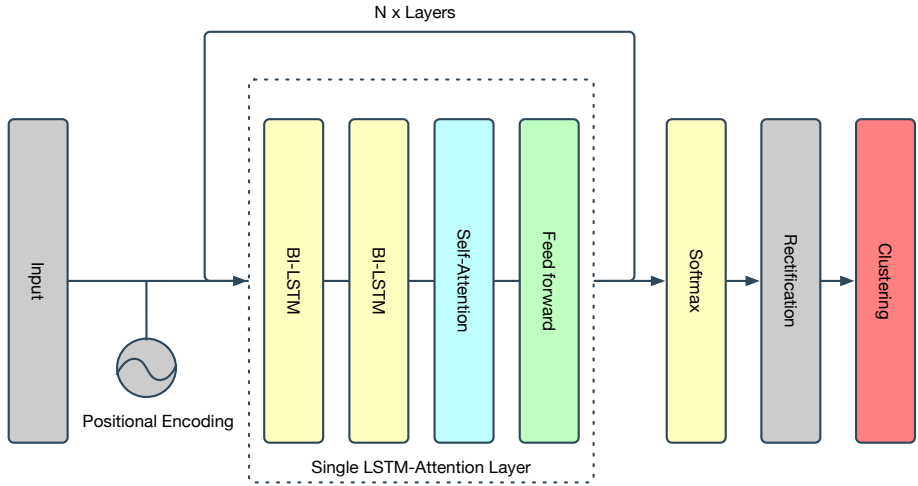

Figure 2: An overview of our model architecture.

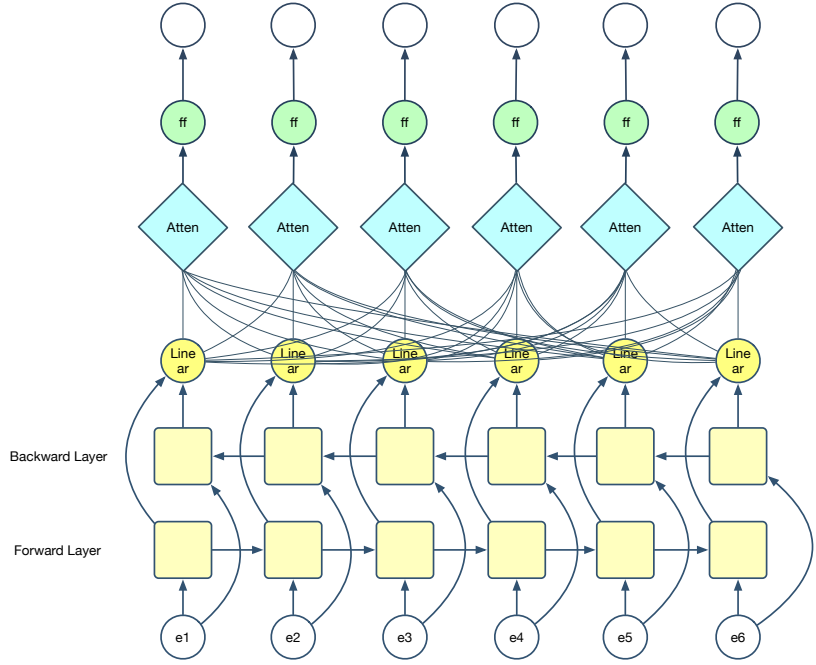

Figure 3: Detail of a single LSTM-Attention layer.

### 3.4 Loss

At the end of this model, I use a Multi-Label Cross entropy loss to do training. Assuming characters $\{x1, x2, x3\}$ can be combined to a new word "$x1x2x3$", then 3 labels would be $\{x1x2x3, x1x2x3, x1x2x3\}$, that is to say all these 3 single characters map to a same word.

Suppose vocabulary size is $s$, $Out_{atten}$ will be mapped to dimension $d_s$ from d via a linear projection:

$$logit_s = W_{ds} * Out_{atten} + b_s$$

Then with a cross entropy loss:

$$loss(logit, class) = -log\left(\frac{exp(logit[class])}{\sum_0^S exp(logit[j])}\right)$$

$$= -logit[class] + log\left(\sum_j exp(logit[j])\right)$$

I use Adam algorithm to train this model with a learning rate of 0.0001. In experiments, learning rate like 0.001 or bigger will cause residual to vibrate.

### 3.5 Rectification

After training, I take $n$ labels with $n$ highest probabilities of each word. I call this the original character's "imaginations". Assuming K a Hyper-parameter:

$$Ima_K = topK(logit_s, K)$$

Given the sentence, we can know which imagination match that very position. So we can choose every character a label which is highly likely that position's mapping word by probabilities and word-match.

We can find these words' embedding in the vocabulary, and feed them to a clustering model. Meanwhile, the "imagined" word does not need to be the real word, characters still can be grouped if they "imagines" the same one. For example, five characters in "2000 年" can be grouped if they all reflect to "2008 年".

### 3.6 Clustering

Cluster number is a hard to decide parameter. So we can give cluster number an approximate value: Ideally, every character in one word should map to that word, so we can count the number of different labels on all positions in a sentence as the value. Clustering model chosen is another issue. Models with best performances in tests are as below:

1. Gaussian Mixture model(GMM). This model randomly chooses some points, and assume each cluster as a single Gaussian distribution. In iterations, model will automatically calculate the expects and variances of all clusters. GMM is accurate and stable in actual experiments. Nevertheless, it will take two to three times long in training than other models.
2. K-Means(MacQueen, 1967) This model first randomly chooses some centroids, and then repeatedly calculate each cluster's inner points and centroids until stable. K-Means is fast and accurate, but sometimes it is unstable so to lose some accuracies in my case.

Currently, I choose GMM as clustering model as it is more stable.

## 4 Experiments

ICWB2-data(Emerson, 2005) is a dataset which is jointly released by Peking University, City University of Hong Kong, Taiwan CKIP, Academia Sinica, and Microsoft

Research Institute of China for training CWS models. Among them, PKU and MSR are simplified Chinese data sets, and AS and CITYU are traditional Chinese data sets.

In my tests, CLNN was trained with *_training.utf8, and evaluated with *_test_gold.utf8. None of the word vocabularies were used because this model only supports sentence training. Every metrics of CLNN in the below figures has been tested more than three times.

PKUSEG tool has a pre-trained "news" model online which was trained in MSR data set, so I used this model when evaluating MSR. For other data set, the default model was used with the corresponding dictionary.

THULAC tool also has a pre-trained model online, which proved helpless to improve accuracy in these data sets. So I used the corresponding dictionary for every data set.

CLNN has some Hyper-parameters, first of all is batch size. Commonly, a relatively big batch size can reduce over-fitting effectively, here I used 16 – 64 sentences per batch, according to each single data set (and my GPU). Embedding dimension was set to 512, with an expand ratio 4. Learning rate was 0.0001 with Adam algorithm. Model layers was set to 6.

Table 1 shows the results of all three models on MSR data set. We can see without the interference of OOV words, CLNN could have the highest F-score. With OOV words, CLNN had an F-score of 94.2% and the highest IV recall rate of 98.8%, but the precision rate was affected greatly. This phenomenon will be explained in the next chapter.

| Dataset | MSR | | | | | | | |
| --- | --- | --- | --- | --- | --- | --- | --- | --- |
| | No OOV | | | OOV | | | | |
| | Recall | Precision | F-score | Recall | Precision | F-score | IV Recall | OOV Recall |
| CLNN | 99.10% | 99.00% | 99.10% | 96.40% | 92.10% | 94.20% | 98.80% | 6.10% |
| PKUSEG | / | / | / | 93.60% | 95.70% | 94.70% | 94.20% | 85.20% |
| THULAC | / | / | / | 92.40% | 93.10% | 92.80% | 93.70% | 73.10% |

Table 1: Different models' performances on MSR (F-score, %).

Table 2 shows results on PKU data set. THULAC kept a great performance, and

CLNN performed well in tests without OOV words, but was generally in experiments with OOV words.

| Dataset | PKU | | | | | | | |
|---|---|---|---|---|---|---|---|---|
| | No OOV | | | OOV | | | | |
| | Recall | Precision | F-score | Recall | Precision | F-score | IV Recall | OOV Recall |
| CLNN | 95.90% | 97.60% | 96.80% | 90.50% | 86.40% | 88.40% | 95.40% | 23.90% |
| PKUSEG | / | / | / | 89.20% | 86.50% | 87.80% | 93.30% | 38.20% |
| THULAC | / | / | / | 92.90% | 95.10% | 94.00% | 93.80% | 82.60% |

Table 2: Different models' performances on PKU (F-score, %).

Table 3 shows the test result on CITYU data set in traditional characters. PKUSEG and THULAC models do not support traditional characters, so no comparison in this figure.

| Dataset | CITYU | | | | | | | |
|---|---|---|---|---|---|---|---|---|
| | No OOV | | | OOV | | | | |
| | Recall | Precision | F-score | Recall | Precision | F-score | IV Recall | OOV Recall |
| CLNN | 99.00% | 99.00% | 99.00% | 93.10% | 86.30% | 89.50% | 98.60% | 7.40% |

Table 3: Different models' performances on CITYU (F-score, %).

As we can see, CLNN can have the highest IV Recall rates on these three data sets and reach the same level of F-score with PKUSEG on MSR.

## 5  Error analyses

OOV words' affections are predictable. CRF-based models could have a well OOV recall rate because a character's position in words it forms is relatively fixed. For example, "国" always shows on an end position. Assuming "法国" never shows in the vocabulary, but because "国" often appears as an end word, "法国" can be recognized as a word. CLNN recognizes words when each character in a word "imagines" a same word. So if "法国" never appears, "法" and "国" will be split, this directly causes precision's reduction.

Several solutions can be used to solve this issue. We can change the mapped words. For example, "1999 年" and "2008 年" will not map to "1999 年" and "2008 年", they all match to a symbol "<year>" in vocabulary. Every proper noun can be translated in this way, like names, times, or places. This will effectively reduce the vocabulary size and improve performances, but this conflicts with my original opinion: discard every pre-defined concept. Secondly, increase the data set. High IV recall rate means this model can effectively deal with IV words, so we can include most regular words and sentences into training set, and improve performance for every specific model with fine-tuning.

In future research, I will focus on resolving this issue elegantly.

Why Self-Attention with BI-LSTM

Theoretically, combinations of different structures will contribute to extract various features. BI-LSTM is able to collect features of neighborhoods; this is just what we want. In experiments of PKU data set, Single eight-level attention model will over-fit and residual will stop at 1.3. However, changed to a four-level BI-LSTM-Attention structure will reduce the residual to 0.6 or less, and improve the accuracy like 3%.

# 6 Conclusion

In this article, I propose a new clustering neural network to do Chinese word segmentation. I combine BI-LSTM with self-attention, then followed by a clustering model to do segmentation.

I am happy to discover a new way, which may resolve the "discrete combination" issue. However, several disadvantages still exist: It cannot make a good use of a single vocabulary, instead it must be trained with sentences. And without probability constraints, some irrational combinations may happen. These issues are to be resolved in the future research.